**Full Title**

Robust and generalizable embryo selection based on artificial intelligence and time-lapse image sequences

**Authors**

J. Berntsen[1,*], J. Rimestad[1], J. T. Lassen[1], D. Tran[2], M. F. Kragh[1,3]

[1] Vitrolife A/S, Jens Juuls Vej 20, 8260 Viby J, Denmark

[2] Harrison AI, Level 21, 60 Margaret St, Sydney NSW 2000, Australia

[3] Department of Electrical and Computer Engineering, Aarhus University, Finlandsgade 22, 8200 Aarhus N, Denmark

*Corresponding author

E-mail address: jberntsen@vitrolife.com (J. Berntsen).

# Abstract

Assessing and selecting the most viable embryos for transfer is an essential part of in vitro fertilization (IVF). In recent years, several approaches have been made to improve and automate the procedure using artificial intelligence (AI) and deep learning. Based on images of embryos with known implantation data (KID), AI models have been trained to automatically score embryos related to their chance of achieving a successful implantation. However, as of now, only limited research has been conducted to evaluate how embryo selection models generalize to new clinics and how they perform in subgroup analyses across various conditions. In this paper, we investigate how a deep learning-based embryo selection model using only time-lapse image sequences performs across different patient ages and clinical conditions, and how it correlates with traditional morphokinetic parameters. The model was trained and evaluated based on a large dataset from 18 IVF centers consisting of 115,832 embryos, of which 14,644 embryos were transferred KID embryos. In an independent test set, the AI model sorted KID embryos with an area under the curve (AUC) of a receiver operating characteristic curve of 0.67 and all embryos with an AUC of 0.95. A clinic hold-out test showed that the model generalized to new clinics with an AUC range of 0.60-0.75 for KID embryos. Across different subgroups of age, insemination method, incubation time, and transfer protocol, the AUC ranged between 0.63 and 0.69. Furthermore, model predictions correlated positively with blastocyst grading and negatively with direct cleavages. The fully automated iDAScore v1.0 model was shown to perform at least as good as a state-of-the-art manual embryo selection model. Moreover, full automatization of embryo scoring implies fewer manual evaluations and eliminates biases due to inter- and intraobserver variation.

# Introduction

Embryo assessment to predict the most viable embryo for transfer has been a challenge since the start of in vitro fertilization (IVF). The introduction of time-lapse in clinical routines (1) has enabled continuous monitoring of embryo development in vitro. This has allowed for the detection of morphological changes and events with the exact time-point of occurrence (2). Based on these morphokinetic parameters, several embryo selection models have been developed (3–7). The end-points of these models were both blastocyst prediction (8,9), genetic status (10–12), gestational sacs (4) and live birth (13–15). Studies have shown a general improvement when time-lapse and morphokinetic selection are used compared with standard incubation (16,17) while other studies have found a need for in-house validation of models before use (18).

During the past decade, a drastic improvement in different technologies used in the IVF labs has been observed, which has implied an increase in workload (19). Even though morphokinetic analysis facilitates a flexible workflow, improved assessment consistency and reduce inter- and intra-observer variations (20–23), there is still considerable challenges to manage embryo selection in a busy lab. This has led to the development of automatic scoring of morphology and morphokinetics. Initial approaches were based on traditional computer vision technology (24,25), while recent methods have used more powerful deep learning frameworks (26–31). The outputs from these methods are used as inputs for decision support systems for ranking or selecting embryos for transfer or cryopreservation.

To completely skip the intermediate steps of estimation of morphology and/or morphokinetics, several newly published methods directly estimate the outcome of an embryo transfer. This can be done based on single images using traditional computer vision technologies (32) or more advanced deep learning technologies (33,34). Covering the whole dynamic embryo development, time-lapse sequences have been used to predict pregnancy in terms of fetal heartbeat (FH) using deep learning (35). This fully automated method obtained an area under the curve (AUC) of the receiver operating characteristic (ROC) of 0.93 for sorting the whole cohort of available embryos in relation to FH.

As deep learning models are typically overparameterized, they easily overfit to the training data set. Therefore, emphasis must be placed on obtaining and evaluating generalization ability across different clinical practices in order to ensure a reliable and trustworthy model that is safe to use in new clinics. To ensure this, a model must be developed and evaluated using very large data sets that cover a range of different clinical practices, incubation conditions, insemination methods and patient cohorts.

In general, deep learning methods can be considered as "black boxes" as interpretation is not transparent. Thus, it has been discussed if biological justification is required for acceptance of computer-generated algorithms to select embryos based on machine-learned combinations of parameters (36). Based on the above, the current analysis aims to develop a fully automated embryo selection model based on a deep learning network that is generalizable, robust and can be supported by biological evidence.

# Materials and methods

## Description of data

Retrospective data from 18 clinics worldwide from between 2011 and 2019 were included in the investigation. No specific methods were applied in the selection of the clinics. Table 1 lists the total number of embryos, the number of transferred embryos and the average female age for each clinic. All data used in the studies were retrospective and provided in a de-identified format. In Denmark, the study described was deemed exempt from notification to the National Committee on Health Research Ethics according to the Act on Research Ethics Review of Health Research Projects (consolidation act no. 1338 of September 1, 2020).

**Table 1. Distribution of total number of embryos, fresh embryos and thawed embryos with known outcome and average female age.**

| Clinic | Total number of embryos | Fresh KID embryos | Thawed KID embryos | Mean age |
|---|---|---|---|---|
| 1 | 24360 | 960 | 21 | 32.0 |
| 2 | 18990 | 1182 | 1490 | 37.2 |
| 3 | 13165 | 550 | 777 | 36.9 |
| 4 | 12640 | 536 | 166 | 32.8 |
| 5 | 10138 | 833 | - | 36.6 |
| 6 | 7679 | 65 | 2768 | 41.3 |
| 7 | 5773 | 386 | 509 | 37.6 |
| 8 | 5757 | 526 | 451 | 36.4 |
| 9 | 4215 | 373 | 242 | 36.4 |
| 10 | 3316 | 455 | 429 | 37.2 |
| 11 | 2729 | 406 | 185 | 34.8 |
| 12 | 1864 | 228 | 152 | 35.6 |
| 13 | 1098 | 140 | 52 | 31.8 |

| 14 | 1042 | 76 | 114 | 36.2 |
| 15 | 1007 | 89 | 113 | - |
| 16 | 817 | 100 | 69 | 35.9 |
| 17 | 759 | 72 | 54 | 36.0 |
| 18 | 483 | 71 | 4 | 36.6 |
| total | 115832 | 7048 | 7596 | 36.0 |

All embryos were cultured in the EmbryoScope incubators (Vitrolife A/S, Aarhus, Denmark). Image sequences were acquired using the settings in each clinic. The number of focal planes varied from 3 to 11. The average time between image acquisition was 15 and 11 minutes for embryos incubated in EmbryoScope-D™ and EmbryoScope Plus™, respectively.

The average patient age was 35.6 with a range from 18 to 52 years. No exclusion of patients was performed. All embryos inserted into the EmbryoScope incubators before 24 hours post insemination (hpi) and removed after 108 hpi were included. Thus, fresh oocytes, cryopreserved oocytes and donated oocytes were all included. Embryos that underwent assisted hatching and biopsy for preimplantation genetic testing (PGT) were also included. On average, each cycle contained 5.7 embryos with a standard deviation of 4.6.

Embryos were classified as FH+, FH-, Unknown or Pending according to the procedure used for the IVY model (35):

- FH+ is when the number of FHs observed on ultrasound after 7 weeks was equal to the number of embryos transferred.
- FH- is when no pregnancy occurred, no FH was observed on ultrasound or the embryos were discarded manually by the embryologist.
- Unknown is when the number of transferred embryos was greater than the number of FHs observed on ultrasound.

- Pending is when the embryo was cryopreserved and no outcome was recorded yet.

In total, 14,644 embryos were transferred (fresh or cryopreserved), resulting in 4,337 FH+ and 10,307 FH- outcomes. In addition, 101,188 embryos were labelled as FH- due to PGT results or manual deselection by embryologists. 23,002 embryos were classified as pending, and 754 embryos were classified as unknown and were not included in the data set. The final data set consisted of 115,832 embryos labelled as either FH+ or FH-. Unknown and pending embryos were not included in the final data set. The data set was randomly split into a training data set (85%) and an independent test data set (15%). This was done for individual embryos across all treatments and clinics.

## Deep learning model

An AI model was trained for binary classification of the positive and negative FHs for the 98,583 embryos in the training data set. Python 3.6.5 and TensorFlow v2.0 were used for the model development. Training was performed using two Nvidia Quadro RTX8000 GPUs.

The overall deep learning network structure for prediction of FH is shown in Fig 1. The inflated 3D (I3D) convolutional neural network (37), a state-of-the-art network for video action recognition, was the basis of the network. The default width of the I3D network was reduced to 25% everywhere. The output from the I3D network was pooled using spatial maximum and average pooling and then concatenated. The output was fed to a bidirectional LSTM (38) with 128 units in each direction resulting in 256 outputs. These were fed to two separate fully connected layers for binary classification of FH and the auxiliary output discard, both with sigmoid activation functions. The model was trained using the Adam optimizer (39) and the one-cycle-policy (40) with an initial learning rate of 1e-5 and a maximum learning rate of 1e-4. Furthermore, a batch size of 64 and a dropout of 25% were used. Hyperparameter estimation and initial experiments were performed with 5-fold cross-validation on the training data. The final model was trained on the whole training data set without a validation data set (41). The final training used the hyperparameter settings found during the 5-fold cross-validation.

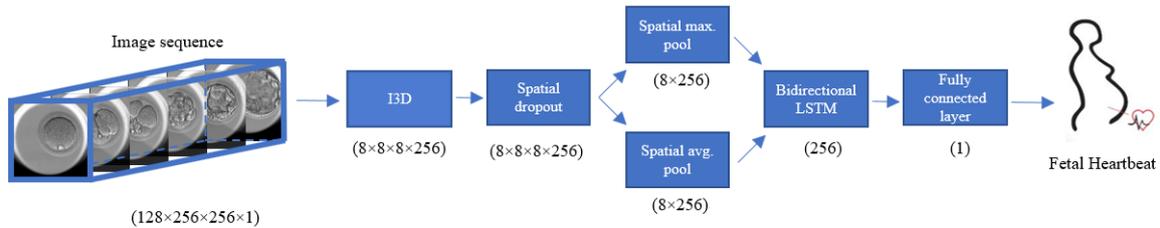

**Fig 1. Overall architecture of the deep learning network used for prediction of fetal heartbeat.** The image sequence is fed to a 3D CNN based on the I3D architecture. The output from the CNN is reduced in dimensions and fed to an LSTM network to utilize temporal information. Finally, a fully-connected layer predicts the fetal heartbeat. The numbers in the figure show the output dimensions from each step in the sequence.

The input to the model was 128 frames sampled one hour apart with one focal plane and a resolution of 256x256 pixels. The sequence was offset by 12 hpi, thus effectively covering 12-140 hpi. During training, a single batch was sampled with 50% from FH+ KID embryos, 10% from FH- KID embryos and 40% from discarded embryos. Input data were augmented in two stages. First, the whole sequence was offset by a random time in [-0.5, 0.5] hours, the focal plane was sampled at –1, 0 and +1 from the central focal plane with a probability of 25%, 50% and 25%, respectively, jitter was introduced by random time shifts of +/- 0.375 hours, and the end time of the sequence was randomly truncated to between 108 and 140 hpi. The sequence was finally padded with zeros to always include 128 frames. Next, the sampled data sequence was applied temporal coherent random cut out (42), 90-110% rescaling, up to 10% translation horizontally and vertically, brightness, horizontal flipping and up to 10-degree rotation clockwise and counter clockwise.

Training was done for 9,264 mini batches, corresponding to 80 epochs for FH+ embryos. The model predicted both FH and discard. For both outputs, the focal loss function (43) with a gamma of 2.0 and an alpha of 0.5 was used. Finally, after training, the FH output from the model was rescaled from [0, 1] to [1.0, 9.9].

## Clinic hold-out test

To evaluate how the model generalizes to new clinics, a leave-one-clinic-out cross-validation was performed on the 12 clinics that each had more than 250 KID embryos. That is, 12 unique models were trained on the remaining clinics (17 clinics in total). The training was performed using the same settings. The trained model was then evaluated on all data from the test clinic.

## Morphokinetic annotations

In some of the clinics, embryo morphology and morphokinetic events were recorded. Annotations were performed according to guidelines by Vitrolife (44) and published nomenclature (2). The following annotations were used: number of pronuclei (PN), time of PN fading (tPNf), cleavage to 2-blastomeres (t2), 3-blastomeres (t3) and 5-blastomeres (t5), time to full blastocyst (tB) and grading of inner cell mass (ICM) and trophectoderm (TE).

Direct cleavages were analysed by defining a direct 1 to 3 division (DC1-3) when t3-tPNf < 5h and a direct 2 to 5 division (DC2-5) when t5-t3 < 5h. If DC1-3 or DC2-5 was not observed, the embryo was defined as having no direct cleavages (no-DC).

For cycles with enough morphokinetic annotations, the KIDScore D5 v3 (45) score was calculated. The model requires annotations of PNs, t2, t3, t5, tB, ICM and TE.

## Statistical analysis

R software version 3.5.3 was used for statistical analysis. For the evaluation of classification performance, the pROC-package (46) was used for estimating ROC curves and calculating the AUC. Tests for significant differences between AUCs were carried out using either paired or unpaired DeLong's one-tailed test. The Mann–Whitney–Wilcoxon test was used to test differences in morphokinetic variables between different groups. A $p < 0.05$ was considered significant.

# Results

All subsequent analyses were performed on the independent test data set that was not used during training. Totally, this data set contained 17,249 embryos of which 2,212 were transferred embryos with known outcome (KID embryos). The overall model performance was evaluated for its sorting capability using AUC. For KID embryos, the AUC was 0.67 with a 95% confidence interval of 0.64-0.69 (Table 2). If the whole cohort was considered, the AUC was 0.95 with a 95% confidence interval of 0.95-0.96.

**Table 2. Number of KID embryos, AUC and 95% confidence intervals for the following sub-groups: age, insemination method, length of incubation and fresh vs cryopreserved embryo transfer.**

| Parameters | Sub-group | n | AUC | 95% C.I. |
|---|---|---|---|---|
| **Overall** | | 2212 | 0.67 | 0.64 - 0.69 |
| **Age** | <30 | 177 | 0.69 | 0.61 - 0.77 |
| | 30-34 | 444 | 0.63 | 0.58 - 0.68 |
| | 35-39 | 655 | 0.67 | 0.63 - 0.72 |
| | >39 | 598 | 0.66 | 0.61 - 0.72 |
| **Insemination method** | ICSI | 738 | 0.69 | 0.65 - 0.73 |
| | IVF | 343 | 0.67 | 0.62 - 0.73 |
| **Length of incubation** | D5 | 1894 | 0.65 | 0.63 - 0.68 |
| | D6 | 303 | 0.66 | 0.57 - 0.74 |
| **Transfer protocol** | Fresh | 1070 | 0.69 | 0.66 - 0.72 |
| | Cryopreserved | 1142 | 0.65* | 0.61 - 0.68 |

A significantly lower AUC for a sub-group compared to the AUC for the remaining sub-groups is indicated with a star ($p < 0.05$).

## Sub-group analysis

The following sub-groups were analysed: patient age (for the groups <30, 30-34, 35-39 and >39 years), insemination method (IVF or ICSI), length of incubation (5 or 6 days) and fresh vs cryopreserved embryo transfer (Table 2). For age groups, the AUC for KID embryos varied between 0.63 and 0.69, with the lowest AUC for the age group 30-34 years. With regards to insemination method, the AUC was 0.69 and 0.67 for ICSI and IVF, respectively. For length of incubation, the AUC was 0.65 and 0.66 for D5 and D6, respectively. Cryopreserved embryos had a significantly lower AUC of 0.65 compared to fresh transfers with an AUC of 0.69.

## Clinic hold-out test

To investigate how the chosen model architecture and training data generalize to new clinics, a clinic hold-out test was performed. The AUCs for the individual clinics varied between 0.60 and 0.75 (Table 3). For clinic 1 and 4, the AUC was significantly smaller than for the rest of the clinics included in the test data.

**Table 3. AUC and 95% Confidence Interval in the clinic hold-out test for KID embryos.**

| Clinic | AUC   | 95% C.I.    |
|--------|-------|-------------|
| 1      | 0.63* | 0.59 - 0.66 |
| 2      | 0.66  | 0.64 - 0.68 |
| 3      | 0.65  | 0.61 - 0.68 |
| 4      | 0.60* | 0.56 - 0.65 |
| 5      | 0.75  | 0.71 - 0.78 |
| 6      | 0.72  | 0.70 - 0.74 |
| 7      | 0.68  | 0.64 - 0.72 |
| 8      | 0.66  | 0.62 - 0.70 |
| 9      | 0.65  | 0.60 - 0.70 |
| 10     | 0.65  | 0.61 - 0.69 |
| 11     | 0.68  | 0.64 - 0.73 |
| 12 | 0.68 | 0.62 - 0.73 |

Only clinics with more than 250 KID embryos are included. A significantly lower AUC for a clinic compared to the AUC for the hold-out data set is indicated with a star ($p < 0.05$).

## Correlation with morphology and morphokinetic annotations

The biological validity of the model was evaluated by correlating the scores with morphological and morphokinetic parameters. This comparison was only done for the embryos in the test data set that had morphological and morphokinetic annotations.

Responses to direct cleavages were tested on the subset of the test data for which the timings tPNf, t3 and t5 were all annotated (Fig 2A). For the whole embryo cohort, the scores were significantly different across all three groups ($p < 0.0001$). For the blastocyst group (i.e. with a tB annotation), the scores in the no-DC group were significantly different from both DC1-3 and DC2-5 ($p < 0.0001$). However, there was no significant difference between DC1-3 and DC2-5 ($p = 0.18$).

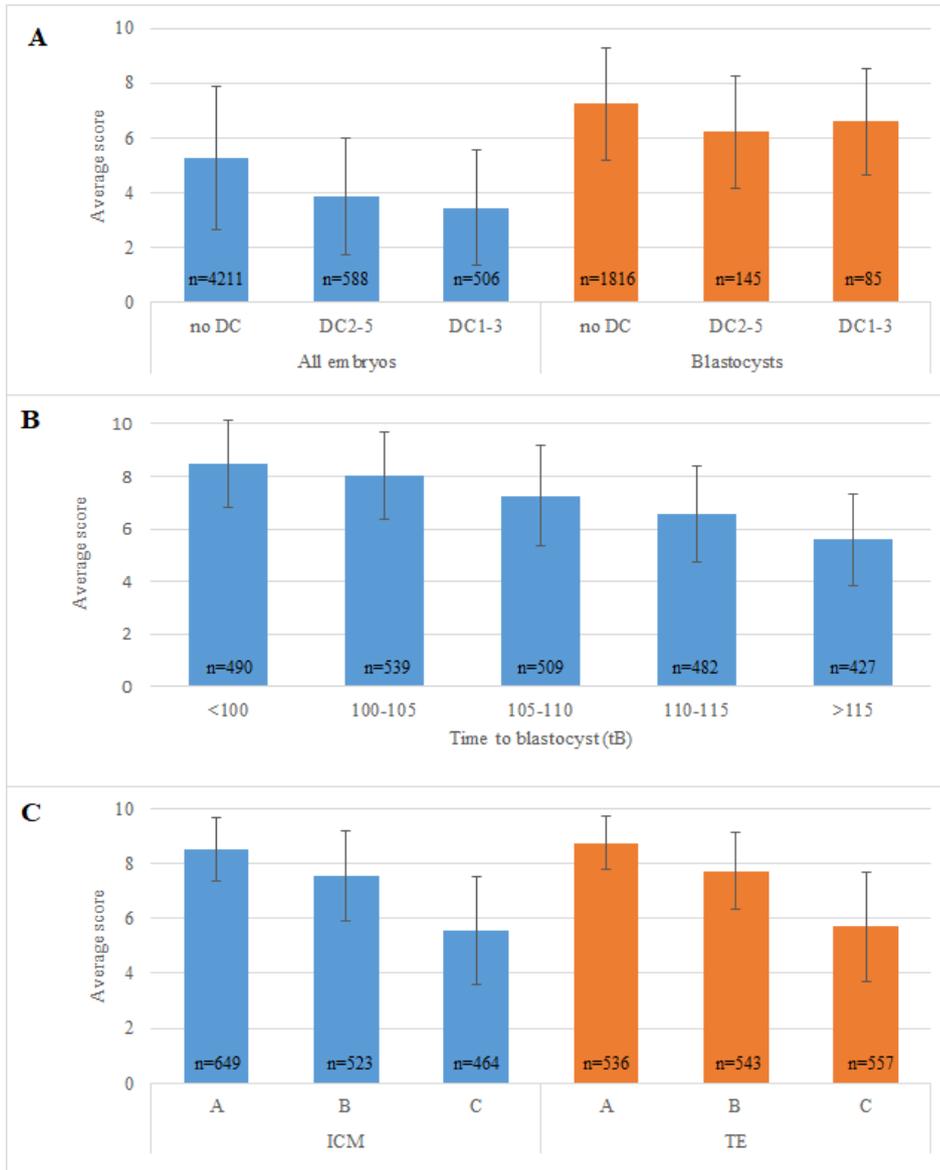

**Fig 2. iDAScore for different morphokinetic groups.** (A) Average score for embryos with no direct cleavages (no DC), direct cleavage from 2 to 5 blastomeres (DC2-5) or direct cleavage from 1 to 3 blastomeres (DC1-3), respectively. Average scores are shown for the whole cohort and for embryos that reached the blastocyst stage. (B) Average score for blastocysts where time to blastocyst was <100, 100-105, 105-110, 110-115 or >115 hpi, respectively. (C) Average score for embryos with ICM and TE grades A, B or C, respectively. The number of embryos in each group is shown inside the bar. Lines show the standard deviation within each group.

Responses to development speed were tested on the subset of the test data for which tB was annotated (Fig 2B). The average score decreased with slower development and there were significant differences between the average scores in all tB groups (p < 0.0001). For embryos with annotated ICM and TE, the average score decreased with lower grades (Fig 2C) and there were significant differences between different grades (p < 0.0001).

## Comparison with the KIDScore D5 v3 model

The performance of the model was compared with the KIDScore D5 v3 model (45) in cycles where enough morphological and morphokinetic parameters were annotated. For KID embryos, only embryos with annotation of tB, ICM and TE were included. Of the total 17,249 embryos in the test set, there were 7,932 annotated embryos, and of the 2,212 KID embryos, there were 1,094 annotated embryos. The ROC curve was calculated for both models and for both the whole cohort and the KID embryos (Fig 3). For sorting the whole cohort, the AUC of 0.92 for the iDAScore v1.0 model was significantly higher than the AUC of 0.89 for KIDScore D5 v3. For sorting of KID embryos, there was no significant differences between the AUC of 0.67 for the iDAScore v1.0 model and 0.66 for KIDScore D5 v3.

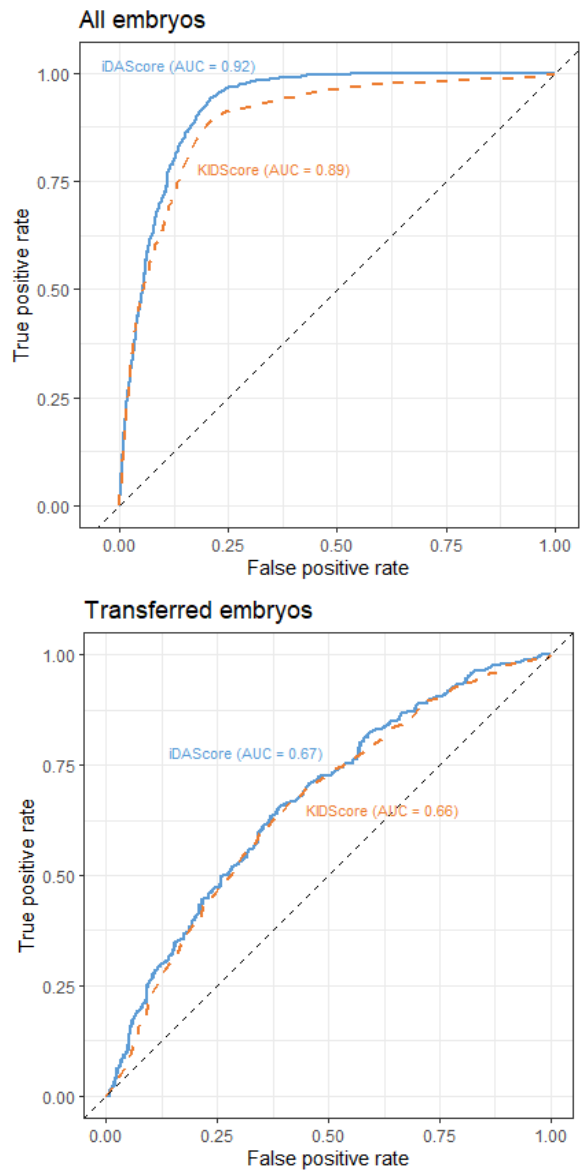

**Fig 3. Comparison between ROC curves based on iDAScore (solid blue) and KIDScore (dotted orange).** The upper plot shows the ROC curve for the whole embryo cohort (n = 7,932), and the lower plot shows the ROC curve for KID embryos (n = 1,094).

# Discussion

The sorting capability for the whole cohort had an AUC of 0.95, which is higher than the 0.93 observed with the IVY model (35). It should be noted that all clinics in the IVY model also contributed with data in this investigation. However, additional new data from these clinics and 5 new clinics were added in this investigation so that the total number of embryos increased from 8,836 to 115,832, the number of embryos with known implantation data (KID) increased from 1,773 to 14,644 and the number of FH+ embryos increased from 694 to 4,337. Thus, the current data set was more than 6 times larger than the data set used for training the IVY model, which means that an even higher performance and robustness is obtained. To our knowledge, this is currently the largest data set used for developing an embryo selection model.

The model's sorting capabilities for KID embryos had an AUC of 0.67 (Table 2) which is lower than the sorting capability for the whole cohort, where the AUC was 0.95. This difference is expected, as it is much more difficult to sort between good morphology blastocysts than to sort within the whole cohort that includes a range from arrested embryos to good morphology blastocysts. Which measure is the most relevant depends on the task of the model. If the purpose is to have a model that is fully automated and sorts between all embryos, the high AUC for sorting the whole cohort is the most relevant measure. However, if the user pre-screens the embryos to find potential transfer candidates, the lower AUC for sorting only within the KID embryos is the most relevant. As both tasks are relevant for embryo selection algorithms and as practical use will depend on the actual clinical setting, we propose that both measurements should be reported in future publications.

## Sub-group analysis

When designing a selection model that is intended to be used across a wide range of clinical settings, patients and culture conditions, it is extremely important to investigate any potential bias within sub-groups in the data set. For nearly all sub-groups, the AUC was within the 95% confidence interval of the AUC of the overall ROC curve. Only the cryopreserved sub-group had a significantly lower AUC than the other sub-groups. The reason for this is probably that other processes (e.g. the vitrification/warming and subsequent endometrium preparation) have an impact on a successful implantation. Thus, the sorting becomes more difficult, which results in a lower AUC.

To our knowledge, no other selection models have been tested on different sub-groups within a large independent data set. However, it is very important to check that models perform comparable on different sub-sets and do not exhibit general biases. When a selection model uses age as a parameter, testing on age sub-groups becomes even more important. The inclusion of age will inevitably improve the overall model performance significantly, as age is one of the best predictors for successful implantation. However, including age may not improve the sorting capability on treatment level (i.e. the embryo cohort of a single patient), which ultimately is the essence of an embryo selection model. Both the STORK algorithm (28) and AIR E (32) include age as input in their models. However, as no age sub-group analysis was performed, the sorting capabilities of these models on treatment level remain to be documented.

## Clinic hold-out test

Several studies have shown the need for testing selection models on in-house data before clinical use (18,47). In this analysis, we used a clinical hold-out validation approach where the model was first trained on data excluding a specific clinic and then tested on this clinic. As shown in Table 3, these models showed a similar sorting capability with the majority being within the 95% confidence interval of the final model. Clinic 1 and 4 with young women (Table 1) had a significantly lower AUC than on the test set. For the younger women in these clinics, the majority of the transferred KID embryos were probably top-quality blastocysts. For the older women in the other clinics, it is likely that a more diverse cohort of KID embryos are transferred. As it is more difficult to sort within a group of homogeneous top-quality blastocysts than within a more diverse group of KID embryos, the AUCs were lower for clinics 1 and 4.

It should be noted that a high AUC does not imply a high implantation rate or vice versa. A high AUC specifically refers to how a model sorts embryos within a given cohort. Thus, differences in AUCs can result from poor model generalization, but indeed also from differences in patient cohort and clinical practices as above.

## Correlation with morphology and morphokinetic annotations

For direct cleavages, Fig 2A shows a clear decrease in the average score for the whole cohort of embryos with direct cleavages annotated. This agrees with several day 3 embryo grading models (4,5) where embryos with direct cleavages are excluded or given a low score. It should also be noted that DC1-3 embryos had a lower score than DC2-5 embryos which agrees with other studies (48) on day 3 transfers. Furthermore, it was observed that for blastocysts, the impact of direct cleavages on the score was smaller than for the whole cohort. This aligns with the hypothesis of "self-correction" where the embryo can correct mitotic cell divisions errors and, if the blastocyst stage is reached, have a relatively better implantation rate compared to a day 3 transfer (48,49).

A clear significant effect was observed between the score and time to blastocyst. This has also been investigated in several other studies (36). The timing and grade of blastocyst expansion have been shown by several investigators to be important predictors of implantation (50,51). In addition, time to blastocyst is also a direct parameter in the KIDScore D5 logistic regression model (45).

For both ICM and TE, there was a clear correlation between the score and the grade. This supports several studies that have shown how ICM and TE grades correlate with implantation (26,51–53).

Thus, the above analyses clearly demonstrate that iDAScore outputs correlate with the most established biologically meaningful parameters that are known to have an impact on implantation. It should be underlined that the model was never trained on any of these parameters, but apparently indirectly learned features that correlate with these biologically meaningful parameters.

## Model comparisons

Currently, there is no general agreement on how to compare the performance of different embryo selection models. Some studies have measured model performance using well-known diagnostic performance parameters such as sensitivity, specificity and accuracy (54,55), while other studies have used AUC to measure the sorting capabilities (4,28,35). Measurements such as sensitivity, specificity and accuracy are excellent for evaluating the overall performance of a diagnostic test. Usually, these models have a binary output: positive or negative. However, in a clinical setting where the task is to rank embryos, we believe that AUC is the best measurement for models that have more than two outputs. The AUC values integrate the sensitivity and specificity over the whole range of possible score values. In that way, an estimate of the overall sorting capability of the model is obtained. In addition, based on a ROC curve (Fig 3) it is possible to estimate the sensitivity and specificity at any score threshold.

The best method for comparing different models is to evaluate AUC on the same independent data set. In the present work, we compared a fully automated model (iDAScore v1.0) based on deep learning with the state-of-the-art KIDScore D5 v3 morphokinetic model based on manual subjective labelling. In this comparison, the deep learning model had a significantly higher AUC for sorting within the whole cohort of embryos (Fig 3). The AUC for KID embryos was numerically higher for the fully automated model, but there was no significant difference. Thus, the fully automated model in general performs better than the more subjective and annotation-based KIDScore D5 v3 model.

Based on the ROC curve for the whole cohort, it was observed that the iDAScore v1.0 and KIDScore D5 v3 perform nearly identical up to a true positive rate of 0.5. This part of the curves is based on the high scoring embryos. Thus, for these embryos, the sorting capability is nearly identical. However, above the true positive rate of 0.5, the ROC curve for iDAScore v1.0 becomes higher than for KIDScore D5 v3. This part of the curve is based on embryos with intermediate scores where iDAScore v1.0 performs significantly better than KIDScore D5 v3. The reason for this is that iDAScore v1.0 was also trained on discarded embryos, while KIDScore D5 v3 was only trained on transferred KID embryos. Thus, if the aim of a model is a fully automated system with the purpose of sorting all embryos in a cycle, it is important that the model learns both how transferred embryos implant and how lower scoring embryos perform. In example, if arrested cleavage stage embryos are not part of the training data, the model's performance on this kind of embryos will be unpredictable as the model has never seen this type of embryos before. This can be obtained by including discarded embryos as in the current investigation.

## Conclusion

In summary, we have developed a fully automated deep learning model (iDAScore v1.0) based on 115,832 embryo time-lapse sequences. The selection model has an AUC of 0.67 for KID embryos. It was demonstrated that this fully automated model performs better than the state-of-the-art morphokinetic KIDScore D5 v3 model. The high performance was obtained without the need for assessment/annotation by the embryologist.

In addition, it was demonstrated that the model was applicable to clinics excluded from the training data. When the model was tested on different stratifications within the test data set, it was also shown that the model can be generalized across age groups, insemination methods and length of incubation. The sorting capability within the different stratifications was close to the overall performance of the final model on the test data set.

Finally, iDAScore v1.0 score correlated with well-established embryo development and morphology parameters. Thus, the model gave lower scores to embryos that had direct cleavages, high scores to faster developing blastocysts and high scores to embryos with high-quality trophectoderm and inner cell mass. Even without training on these parameters, the model has thus indirectly learned that these parameters are important for implantation.

## Acknowledgements

We highly appreciate the collaboration and exchange of data from the following clinics: Aagaard Fertility, Ciconia Fertility Clinic, Complete Fertility Centre Southampton, Fertility Clinic Skive, Horsens Fertility Clinic, IVF Australia-Alexandria, IVF Australia-Canberra, IVF Australia-Greenwich, IVF Australia-Hunter, IVF Australia-Westmead, Kato Ladies Clinic, Maigaard Fertility Clinic, Melbourne IVF, Queensland Fertility Group, SIMS Cork, SIMS Dublin, SIMS Rotunda and TasIVF Hobart.

The authors would also like to acknowledge Csaba Pribenszky, Mark Larman, Markus Montag, Mia Steen Johnsen and Tine Qvistgaard Kajhøj for their invaluable support and suggestions to the manuscript.